\documentclass{article} 
\usepackage{nips15submit_e,times}
\usepackage{hyperref}
\usepackage{url}

\usepackage{amsmath}	
\usepackage{amsfonts}
\usepackage{wrapfig}
\usepackage{amssymb}
\usepackage{amsthm}
\usepackage{graphicx}

\usepackage{tabularx}

\usepackage[square, numbers,sort]{natbib}
\usepackage{natbib}
\usepackage{algorithm}
\usepackage{algorithmicx}
\usepackage{algpseudocode}
\usepackage{xparse}


\DeclareDocumentCommand{\foocmd}{ O{default1} O{default2} m }{#1~#2~#3}
\DeclareMathOperator*{\argmin}{arg\,min}

\renewcommand{\vec}[1]{{\bf #1}}

\newtheorem{theorem}{Theorem}
\newtheorem{definition}[theorem]{Definition}
\newtheorem{proposition}[theorem]{Proposition}
\newtheorem{lemma}[theorem]{Lemma}
\newcommand{\thus}{{^\mathrm{th}}}
\newcommand{\stus}{{^\mathrm{st}}}

\newcommand{\vv}{\vec{v}}
\newcommand{\hi}[1][i]{h_{#1}}
\newcommand{\vj}[1][j]{v_{#1}}
\newcommand{\xj}[1][j]{x_{#1}}
\newcommand{\xs}{{\cal X}}

\newcommand{\wij}[1][ij]{W_{#1}}

\newcommand{\nn}[1][k]{{N_{#1}}}
\newcommand{\nov}{{N_\mathrm{vis}}}
\newcommand{\noh}{{N_\mathrm{hid}}}

\newcommand{\eres}{E_\mathrm{res}}

\DeclareDocumentCommand{\wijkl}{ O{i} O{j} O{k} O{l} }{\wij[#1,#2]^{(#3,#4)}}
\DeclareDocumentCommand{\xik}{ O{i} O{k} }{x_{#1}^{(#2)}}
\DeclareDocumentCommand{\bik}{ O{i} O{k} }{b_{#1}^{(#2)}}
\DeclareDocumentCommand{\hik}{ O{i} O{k} }{h_{#1}^{(#2)}}
\DeclareDocumentCommand{\xvk}{ O{k} }{\vec{x}^{(#1)}}
\DeclareDocumentCommand{\hvk}{ O{k} }{\vec{h}^{(#1)}}
\DeclareDocumentCommand{\bvk}{ O{k} }{\vec{b}^{(#1)}}

\DeclareDocumentCommand{\wkl}{ O{k} O{l} }{w^{(#1,#2)}}
\DeclareDocumentCommand{\hwkl}{ O{k} O{l} }{\hat{w}^{(#1,#2)}}
\DeclareDocumentCommand{\bk}{ O{k} }{b^{(#1)}}
\DeclareDocumentCommand{\xk}{ O{k} }{x^{(#1)}}
\DeclareDocumentCommand{\hk}{ O{k} }{h^{(#1)}}

\DeclareDocumentCommand{\hbk}{ O{k} }{\hat{b}^{#1}}

\newcommand{\Sl}[1][L]{S(#1)}
\DeclareDocumentCommand{\Slc}{ O{1} O{L} }{S_{\xk[L]=#1}(#2)}

\newcommand{\lse}{\log\sum_{\hs}\exp}

\newcommand{\partf}[1][\para]{Z(#1)}

\newcommand{\binary}{\{0,1\}}

\newcommand{\refeq}[1]{Eq.~(\ref{#1})}
\newcommand{\reffig}[1]{Fig.~\ref{#1}}
\newcommand{\refFig}[1]{Figure~\ref{#1}}

\newcommand{\reftab}[1]{Table~\ref{#1}}

\newcommand{\hs}{{\cal H}}
\newcommand{\hsmin}{{\hat{\hs}(\vv)}}
\newcommand{\qmin}{{\hat{Q}}}

\newcommand{\para}{{\boldsymbol\theta}}
\newcommand{\prbm}{\para^\mathrm{RBM}}
\newcommand{\pdbm}{\para^\mathrm{DBM}}
\newcommand{\pmbdbm}{\para^\mathrm{sDBM}}
\newcommand{\pmbOnedbm}[1][]{\para^\mathrm{gBM#1}}

\newcommand{\ebm}{E}

\newcommand{\fbm}{F}
\newcommand{\hmf}{\hat{F}}
\newcommand{\varf}{{F_\mathrm{MF}}}
\newcommand{\KL}[2]{{\mathrm{KL}\left(#1|#2\right)}}

\newcommand{\fBM}[1][L]{gBM$(#1)$}

\title{Soft-Deep Boltzmann Machines }

\author{
Taichi Kiwaki\\
Graduate School of Engineering \\
The University of Tokyo\\
\texttt{kiwaki@sat.t.u-tokyo.ac.jp}
}

 \nipsfinalcopy 

\begin{document}

\maketitle

\begin{abstract}
We present a layered Boltzmann machine (BM) that can better exploit the advantages of a distributed representation. It is widely believed that deep BMs (DBMs) have far greater representational power than its shallow counterpart, restricted Boltzmann machines (RBMs). However, this expectation on the supremacy of DBMs over RBMs has not ever been validated in a theoretical fashion. In this paper, we provide both theoretical and empirical evidences that the representational power of DBMs can be actually rather limited in taking advantages of distributed representations. We propose an approximate measure for the representational power of a BM regarding to the efficiency of a distributed representation. With this measure, we show a surprising fact that DBMs can make inefficient use of distributed representations. Based on these observations, we propose an alternative BM architecture, which we dub soft-deep BMs (sDBMs). We show that sDBMs can more efficiently exploit the distributed representations in terms of the measure. Experiments demonstrate that sDBMs outperform several state-of-the-art models, including DBMs, in generative tasks on binarized MNIST and Caltech-101 silhouettes. 
\end{abstract}

\section{Introduction}

One aspect behind superior performance of deep architectures is the effective use of distributed representations \cite{Hinton:1986dr, Bengio:2009wib}. 
A representation is said {\it distributed} if it consists of mutually non-exclusive features \cite{Hinton:1986dr}. 
Distributed representations can efficiently model complex functions with enormous number of variations by dividing the input space to a huge number of sub-regions with a combination features \cite{Bengio:2009wib}. 
 

Recent analyses have proven efficient use of distributed representations in deep feed forward networks with rectified linear (ReL) activations \cite{Pascanu:2013ue, Montufar:2014tb}.
Such deep networks model complex input-output relationships by dividing the input space to enormous number of sub-regions, that grow exponentially in the number of parameters. 
Multiple levels of feature representations in deep feed forward networks successfully facilitate efficient reuse of low-level representations,
and deep feed forward networks thus can manage an exponentially greater number of sub-regions than shallow architectures. 

It is interesting to ask whether {\it deep generative models} could attain such a property as deep discriminative models. 
To answer this question, it would be useful to compare restricted Boltzmann machines (RBMs) and deep Boltzmann machine (DBMs). 
RBMs are a shallow generative model with distributed representations \cite{Rumelhart:va}. 
Deep Boltzmann machines (DBM) are a deep extension of RBMs \cite{Salakhutdinov:2009uo}. 
DBMs are commonly expected to have a far greater representational power than RBMs while being relatively easy to be trained compared to RBMs. 
However, the expectation of supremacy of DBMs over RBMs  has not ever been validated in a theoretical fashion.

In this paper, we provide both theoretical and empirical evidences that the representational power of DBMs can actually be rather limited in exploiting the advantages of distributed representations. 
 Our contributions are as follows. First, we propose an approximate measure for the efficiency of distributed representations of BMs inspired by recent analysis on deep feedforward networks \cite{Pascanu:2013ue, Montufar:2014tb}. 
Our measure is the number of linear regions of a piecewise linear function that approximates the free energy function of a BM. 
This measure approximates the number of sub-regions that a BM manages in the visible space. 
We show that the depth does not largely improve the representational power of a DBM in terms of this measure. 
This indicates a surprising fact that DBMs can make inefficient use of distributed representations, despite common expectations.
Second, we propose a superset of DBMs, which we dub soft-deep BMs (sDBMs). 
An sDBM is a layered BM where all the layer pairs are connected with topologically defined regularization. 
Such relaxed connections realize {\em soft} hierarchy as opposed to hard hierarchy of conventional deep networks where only neighboring layers are connected. We show that the number of linear regions of the approximate free energy of an sDBM scales exponential in the number of its layers thus can be as large as that of a general BM and can be exponentially greater than that of an RBM or a DBM. 
Finally, we experimentally demonstrate high generative performance of sDBMs.
sDBMs trained without pretraining outperform state-of-the-art generative models, including DBMs, on two benchmark datasets: MNIST and Caltech-101 silhouettes.


\begin{figure}
 \centering
\begin{minipage}{0.25\textwidth}
 \includegraphics[width=1.0\textwidth]{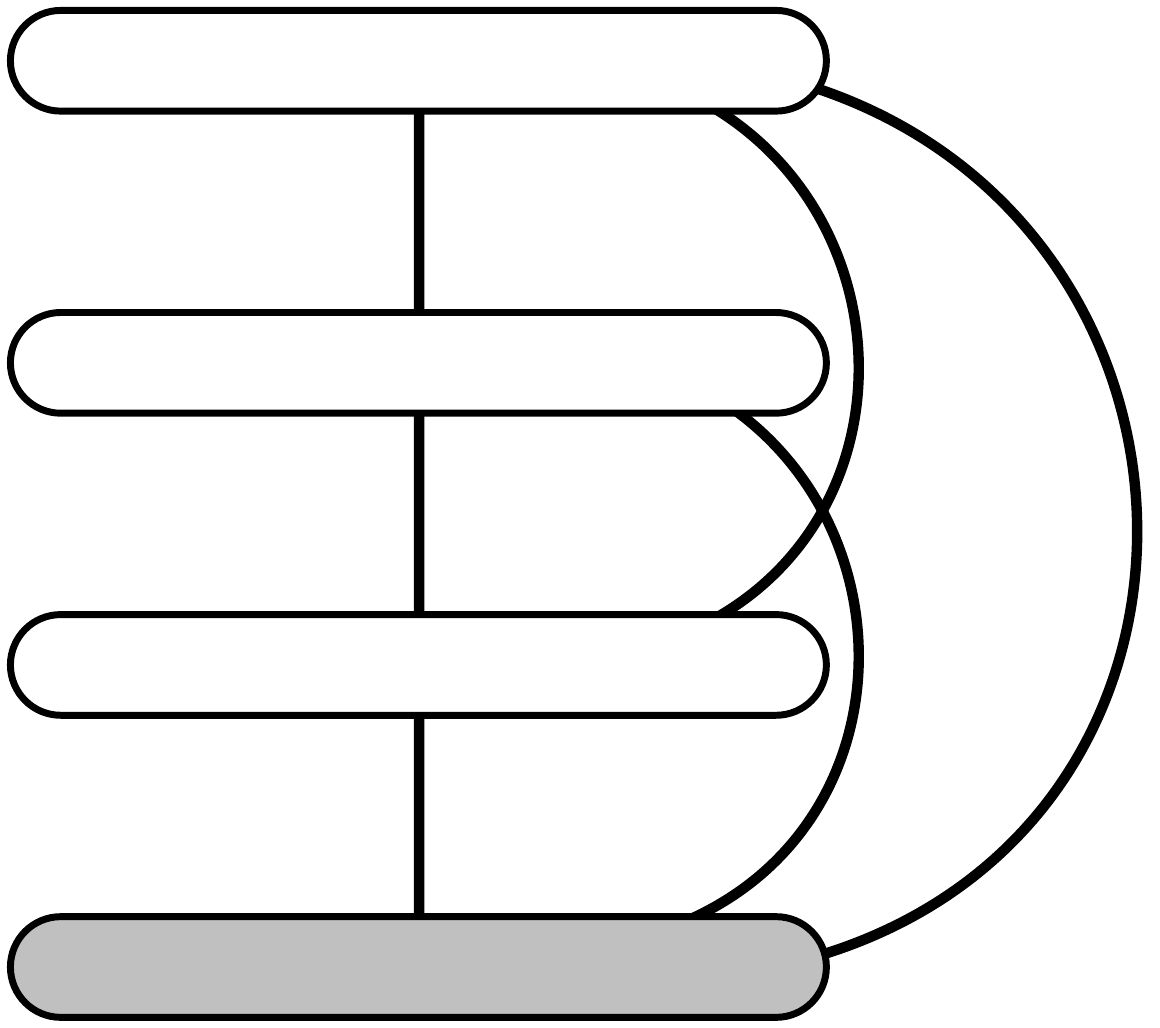}
 \caption{Illustration of an sDBM. All the layer pairs are connected in different magnitudes of strength. }
 \label{sDBM:fig}
\end{minipage}
\hspace{3mm}
 \begin{minipage}{0.67\textwidth}
  \begin{minipage}{0.5\textwidth}
   \includegraphics[width=\textwidth]{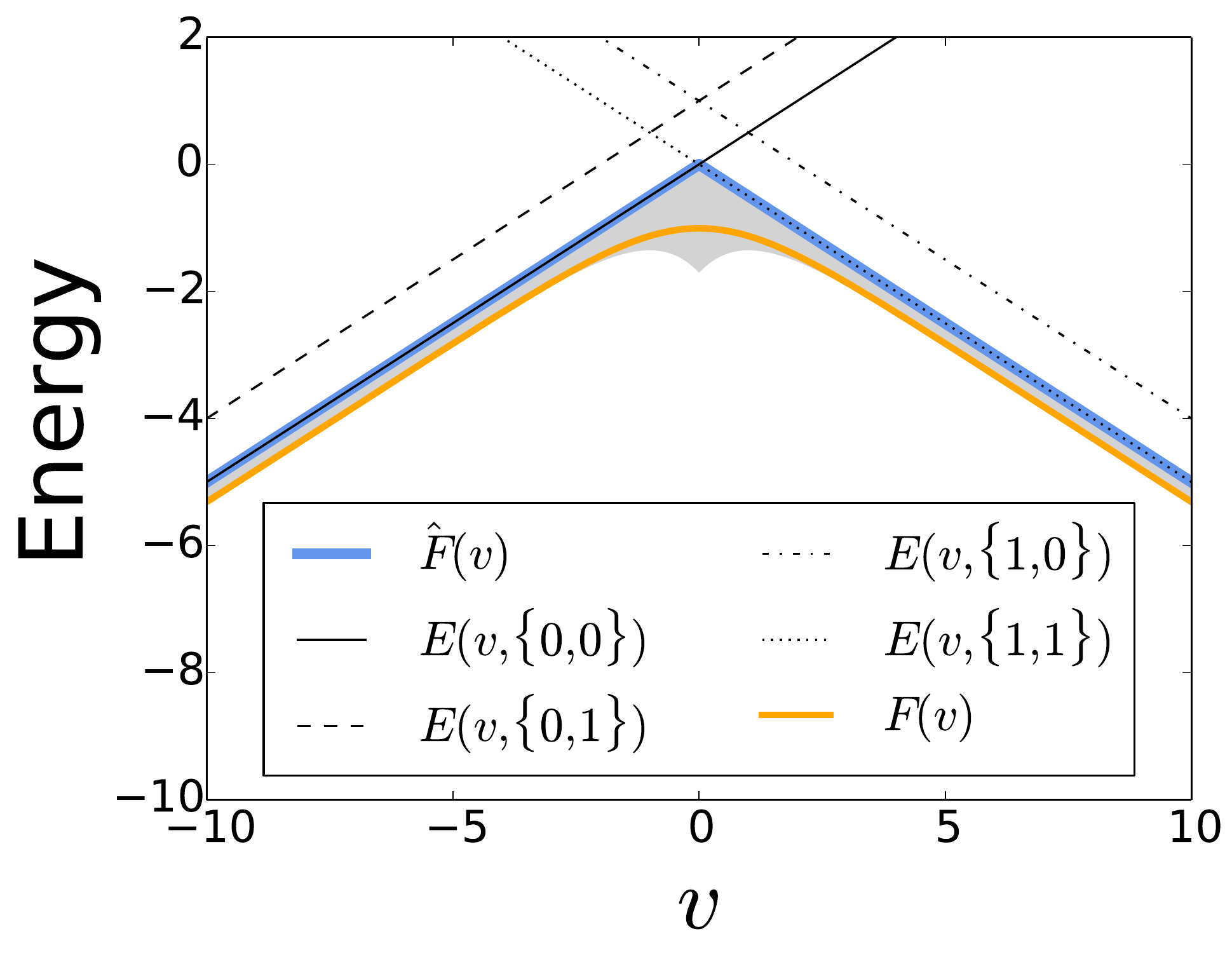}
  \end{minipage}
  \begin{minipage}{0.48\textwidth}
  \includegraphics[width=\textwidth]{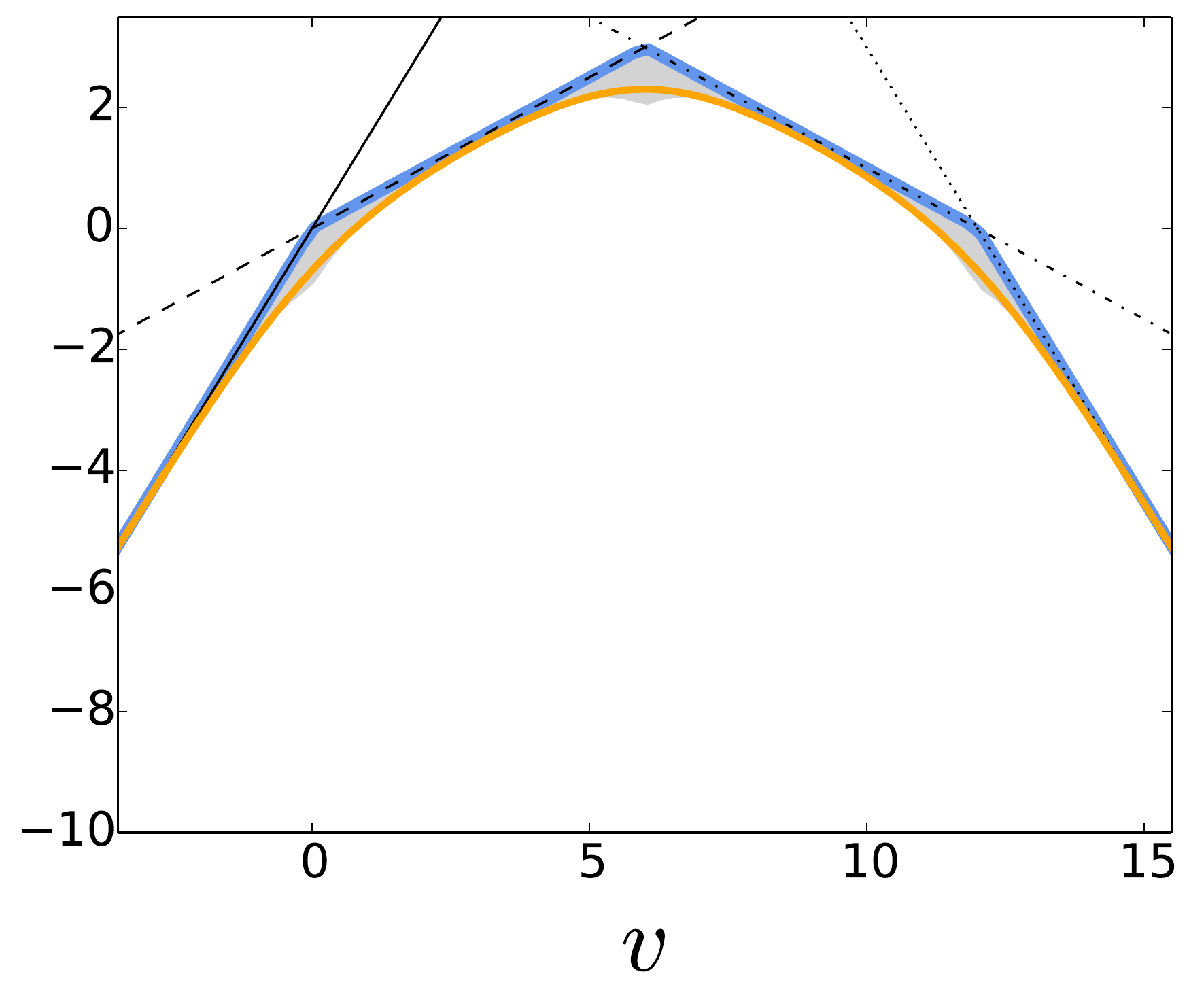}
  \end{minipage}\\
\vspace{-5mm}
 \hspace{2.5cm}{(a)\hspace{4.0cm} (b)}
   \caption{(a) Free energy $F$ and hard-min free energy $\hmf$ of (a) a two-layered DBM and (b) sDBM (i.e., gBM(2)). 
Free energy bounds are indicated with shaded regions. 
All the layers of BMs have only one unit. 
 The sDBM parameters are generated with Algorithm~\ref{mb1DBM:alg} and rescaling.
(best view in color)}
\label{dBM_fe:fig}
 \end{minipage}\\
\end{figure}

\section{Soft-Deep BMs} 
We propose a soft-deep BM (sDBM): a Boltzmann Machine (BM) \cite{Hinton:ub} that consists of multiple layers where {\em all the layer pairs are connected} and connections within layers are restricted. \refFig{sDBM:fig} illustrates an sDBM. 
The energy of an sDBM is defined as:
\begin{align}
\ebm(\xs;\pmbdbm) = -\sum_{0\leq l < k \leq L}\sum_{i=1}^{\nn}\sum_{j=1}^{\nn[l]} \xik \wijkl \xik[j][l] - 
\sum_{k=0}^{L}\sum_{i=1}^{\nn[k]} \bik\xik
,\label{mbdbm:eq}
\end{align}
where $\pmbdbm=\{\wijkl, \bik\}$ is a set of parameters, $\xvk=\{\xik\}$ is the state of the $k\thus$ layer, and $\xs=\{\vv,\hs\}$ is the set of all the units with $\hs=\{\hvk\}$ being the set of hidden layers, and $\vv$ being the visible layer. 
We number layers $\xvk$ s.t. $\xvk[0]$ is the visible layer, and $\xvk[k]$ is the $k\thus$ hidden layer.
Let $L$ be the number of layers, $N$ be the total number of units, $\nn$ be the number of units in $k\thus$ layer, $\nov$ be the number of visible units, and $\noh$ be the number of hidden units.
An sDBM assigns probability $p(\xs;\para) \propto \exp(-\ebm(\xs;\para))$ to a configuration $\xs\in\binary^N$. 

RBMs and DBMs are subsets of sDBMs; RBMs are sDBMs of $L=1$, and DBMs are sDBMs where $\wijkl\equiv 0$ for $k-l>1$.

\section{Quantifying the Efficiency of Distributed Representation in BMs}

In the this section, we define an approximate measure for the representational power of a BM based on its free energy function. 
We compare various BMs in terms of this measure and show that sDBMs could attain richer representations than DBMs and RBMs. 

The free energy of a BM is defined as the negative log probability that the network assigns to a visible configuration without normalizing constant:
\begin{align}
\fbm(\vv;\para) &\triangleq - \log\sum_{\hs} \exp(-E(\vv,\hs;\para)), \label{fe:eq}
\end{align}
where $\sum_{\hs}$ denotes summation over all the hidden configurations, and $\para$ is the set of parameters. 
We would be able to measure the representational power of a BM with the complexity of the free energy function 
because the free energy function contains all the information on the probability distribution that a BM models.

\subsection{Hard-min Approximation of Free Energy}



We here define a piecewise linear approximation of the free energy function of a BM. 
For RBMs, it is widely known that the free energy function can be well approximated with a piecewise linear function \cite{Martens:2013up}. 
This idea can be extended to general BMs that do not have connections between visible units, which include sDBMs, as follows: 
the operation $\lse$ in \refeq{fe:eq} can be regarded as a relaxed max operation; 
the sum is virtually dominated by the smallest energy, i.e., $\sum_{\hs} \exp(-E(\vv,\hs))\approx \exp(-\min_\hs E(\vv,\hs))$
where $\min_{\hs}$ denotes min operation over all possible hidden configurations. 
The negative logarithm of the sum is thus nearly $\min_\hs E(\vv,\hs)$. 
Based on this observation, we define following approximation of the free energy:
\begin{definition}
\normalfont
{\it Hard-min free energy} $\hmf$ of a BM with parameters $\para$ is 
defined as: 
\begin{align}
 \hmf(\vv;\para) &\triangleq \min_{\hs} E(\vv,\hs;\para). \label{hmfe:eq}
\end{align}
\label{def1:th}
\end{definition}
Note that $\hmf(\vv;\para)$ is a piecewise linear function if the BM does not have connections between visible units because $E(\vv,\hs;\para)$ does not have interactions involving multiple visible units. 

Formally, we can show that $\hmf$ bounds $F$ as: 
\begin{theorem}
 Let $\eres(\vv) = - \log\{\sum_\hs \exp(-E(\vv,\hs))  - \exp(-\hmf(\vv))\}$. Then the free energy $\fbm(\vv)$ is bounded as:
\begin{align}
 \hmf(\vv) - \exp(\hmf(\vv)-\eres(\vv))\leq \fbm(\vv)\leq \varf(\vv) \leq\hmf(\vv), 
\end{align}
where $\varf$ is the mean-filed approximation of the free energy. 
\label{lub:th}
\end{theorem}

The tightness of the bound is determined by the dominance of minimum energy $\hmf$ over the free energy. 
The difference between the upper and the lower bounds becomes fairly tight if $\hmf$ is smaller than $\eres$, the contribution of the non-minimum energies on the free energy. 

Theorem~\ref{lub:th} shows that $\hmf$ is a very rough approximation for the free energy; $\hmf$ is less accurate than mean-field approximation $\varf$. 
Nevertheless, the bound can be tight except points where several energy terms nearly achieve the minimum, e.g., boundaries between linear regions of $\hmf$. \refFig{dBM_fe:fig} demonstrates this idea. 
Therefore, we will be able to roughly measure the complexity of the free energy of a BM through quantifying the complexity of $\hmf$. 



A natural way to quantify the complexity of a piecewise linear function is to count the number of its linear regions. 
To quantify the representational power of a deep feedforward network with ReL activation, this strategy was recently applied to the piecewise linear input-output function \cite{Pascanu:2013ue, Montufar:2014tb}.
Inspired by these analyses, we propose to use the number of the linear regions of $\hmf$ to measure a BM's representational power. 
Intuitively, this measure roughly indicates the number of effective Bernoulli mixing components of a BM; 
$\hmf$ with $k$ linear regions will be well approximated by the 
negative probability function of a mixture of $k$ Bernoulli components 
by assigning each component to each region. 
We therefore shall call this measure the number of effective mixtures of a BM:
\begin{definition}
\normalfont
Suppose a BM with no connections between visible units. 
{\it The number of effective mixtures} of the BM is the number of linear regions of the hard-min free energy $\hmf$ of the BM. 
\label{def2:th}
\end{definition}

Obviously from Definitions~\ref{def1:th} and \ref{def2:th}, the maximal number of effective mixtures of a BM is bounded above by the number of its hidden configurations:
\begin{proposition} 
The number of effective mixtures of a BM is upper bounded by $2^\noh$.
\label{gen_ub:th}
\end{proposition}

Note that this proposition tells us nothing about whether this bound is actually achievable by a BM with a certain parameter configuration; we provide positive results in later sections. 

The number of effective mixtures of a BM approximately measures the efficiency of the distributed representation. 
Each configuration of a distributed representation can give rise to a linear region of $\hmf$. 
Therefore, an efficient distributed representation of a BM potentially manages $2^{\noh}$ sub-regions in the visible space. 
The efficiency, however, can substantially be damaged by restricted connections. 

For deep feedforward networks with ReL, \citet{Montufar:2014tb} showed that a deeper network can model a piecewise linear function with much more linear regions than a shallow network with the same number of parameters. The number of the linear regions grows exponentially in the number of the layers. Now we ask a question: is this also true for DBMs in terms of the approximate free energy $\hmf$? Surprisingly, the answer is NO. 
We shall provide proofs in the following sections. 


\subsection{The Number of Effective Mixtures of an RBM}


We first analyze RBMs. 
The free energy function of an RBM can be approximated with a 2-layered feedforward network with ReL \cite{Martens:2013up}. The number of linear regions of the input-output function of such a shallow network has been studied by \citet{Pascanu:2013ue} and \citet{Montufar:2014tb}. With slight modification on their results, we can compute the maximal number of effective mixtures of an RBM:
\begin{theorem}
 The maximal number of effective mixtures of an RBM is $\sum_{j=0}^{\nn[0]} {\nn[1]\choose{j}}$.
\label{nolr_rbm:th}
\end{theorem}
Note that this bound is quite smaller than the upper bound in Proposition~\ref{gen_ub:th} for $\nn[1]>\nn[0]$ because $\sum_{j=0}^{\nn[0]} {\nn[1]\choose{j}} = \Theta({\nn[1]}^{\nn[0]})\ll 2^{\nn[1]}$.

\subsection{The Number of Effective Mixtures of a DBM}

Next we analyze DBMs. 
Here we provide lower and upper bounds on the maximal number of effective mixtures of DBMs. 
We have a lower bound because DBMs are a superset of RBMs:
\begin{proposition}
The maximal number of effective mixtures of a DBM is lower bounded by $\sum_{j=0}^{\nn[0]} {\nn[1]\choose{j}}$. 
\label{nolr_lb_dbm:th}
\end{proposition}

A key idea of the proof on an upper bound, which we show in the appendix, is that 
energies associated with a same configuration in the first hidden layer $\hvk[1]$ have an identical gradient in the space of $\vv$. 
For example, $E(v,\hk[1]=\bullet, \hk[2]=0)$ and $E(v,\hk[1]=\bullet, \hk[2]=1)$ have the same gradient i.e., slope in \reffig{dBM_fe:fig}~(a).
This is because 
$\hk[2]$ does not affect the statistics of $v$ given $\hk[1]$.
The number of linear regions of $\hmf$ is therefore bounded by $2^{\nn[1]}=2^{1}$ because one of the energy terms with the same slope become globally smaller than the other energy term, e.g., $E(v,\hk[1]=0, \hk[2]=0)<E(v,\hk[1]=0, \hk[2]=1)$ for any $v$. 
This generalize to any DBMs leading to a natural but somewhat shocking result where the bound only depends on the number of units in the first hidden layer:
\begin{theorem}
The number of effective mixtures of a DBM with any number of hidden layers is upper bounded by $2^{\nn[1]}$. 
\label{nolr_ub_dbm:th}
\end{theorem}
Depth does not largely help the number of effective mixtures of DBMs. 
This suggests that a distributed representation is inefficiently used in a DBM at least in the scope of the approximate free energy $\hmf$. 
From Proposition~\ref{nolr_lb_dbm:th} and Theorem~\ref{nolr_ub_dbm:th}, we can readily show a serious limitation on the number of effective mixture of DBMs:
\begin{proposition}
 The number of effective mixture of a DBM with $\nn[1]>\nn[0]$ never achieves the bound $2^\noh$.
\end{proposition}



\subsection{The Number of Effective Mixtures of an sDBM}
The key to the limited number of effective mixtures of DBMs is the independency between $\vv$ and $\{\hvk[2], \hvk[3], \ldots\}$ given $\hvk[1]$. 
Conversely, if there exists dependency between the visible and the upper hidden layers even given $\hvk[1]$, 
the limitation over the number of effective mixtures will not hold. 
Bypassing connections of sDBMs therefore might improve the number of effective mixtures. 
\refFig{dBM_fe:fig}~(b) demonstrate this idea by showing that $\hmf$ of an sDBM attains $2^{\noh}=2^4$ linear regions with properly chosen parameters. In this section, we refine this idea for general sDBMs. 

\subsubsection{General BMs as Elemental sDBMs}
We first analyze the number of effective mixtures of a general BM with only one visible unit, which can be regarded as an elemental sDBM. 
Let \fBM\ be a general BM with $L$ hidden units and one visible unit whose energy function is defined as:
\begin{align}
 \ebm(\xk[0:L];\pmbOnedbm[(L)]) = -\sum_{0\leq l < k \leq L} \xk \wkl \xk[l] - \sum_{k=0}^{L} \bk\xk,
\label{gBML:eq}
\end{align}
where we defined $v=\xk[0]$, $\hk=\xk$, and $\xk[l:k] \triangleq \{\xk[l],\ldots,\xk[k]\}$ for $0\leq l<k \leq L$. 
Because we regard a \fBM\ as an elemental sDBM with $L$ layers each of which has only one unit, we index units and parameters with superscripts. We may call the $k\thus$ unit of a \fBM\ the {\it $k\thus$ layer} because of the same reason. 
Let $\Sl$ be a set of one dimensional linear functions defined over the visible unit: $\Sl\triangleq\{E(\xk[0],\xk[1:L];\pmbOnedbm[(L)])| \xk[1:L]\in\binary^L \}$. 


\begin{figure}
 \begin{minipage}{0.5\textwidth}
\begin{algorithm}[H]
   \caption{ A recursive construction of \fBM}
   \label{mb1DBM:alg}
\begin{algorithmic}
 \Function{softDeep}{$L$} 
 \If {$L = 0$}
  \State $\wkl\leftarrow 0$ for $0\leq l < k < \infty$
  \State $\bk\leftarrow 0$  for $0\leq  k < \infty$
  \State \Return $\{\wkl\}, \{\bk\}$
 \Else 
  \State $\{\wkl\}, \{\bk\} \leftarrow $ \Call{softDeep}{$L-1$}
  \State $x \leftarrow 2^{L-1}$
  \State $\wkl[L][0] \leftarrow x$
  \State $\bk[L] \leftarrow 0.5x(1-x)$
  \For{$l=1$ {\bfseries to} $L-1$}
   \State $\wkl[L][l] \leftarrow - x\wkl[l][0]$ 
  \EndFor
  \State \Return $\{\wkl\}, \{\bk\}$
 \EndIf
 \EndFunction
\end{algorithmic}
\end{algorithm}  
 \end{minipage}
\begin{minipage}{0.49\textwidth}
 \begin{minipage}{0.49\textwidth}
  \centering
  \hspace{3mm}
  \includegraphics[width=0.8\textwidth]{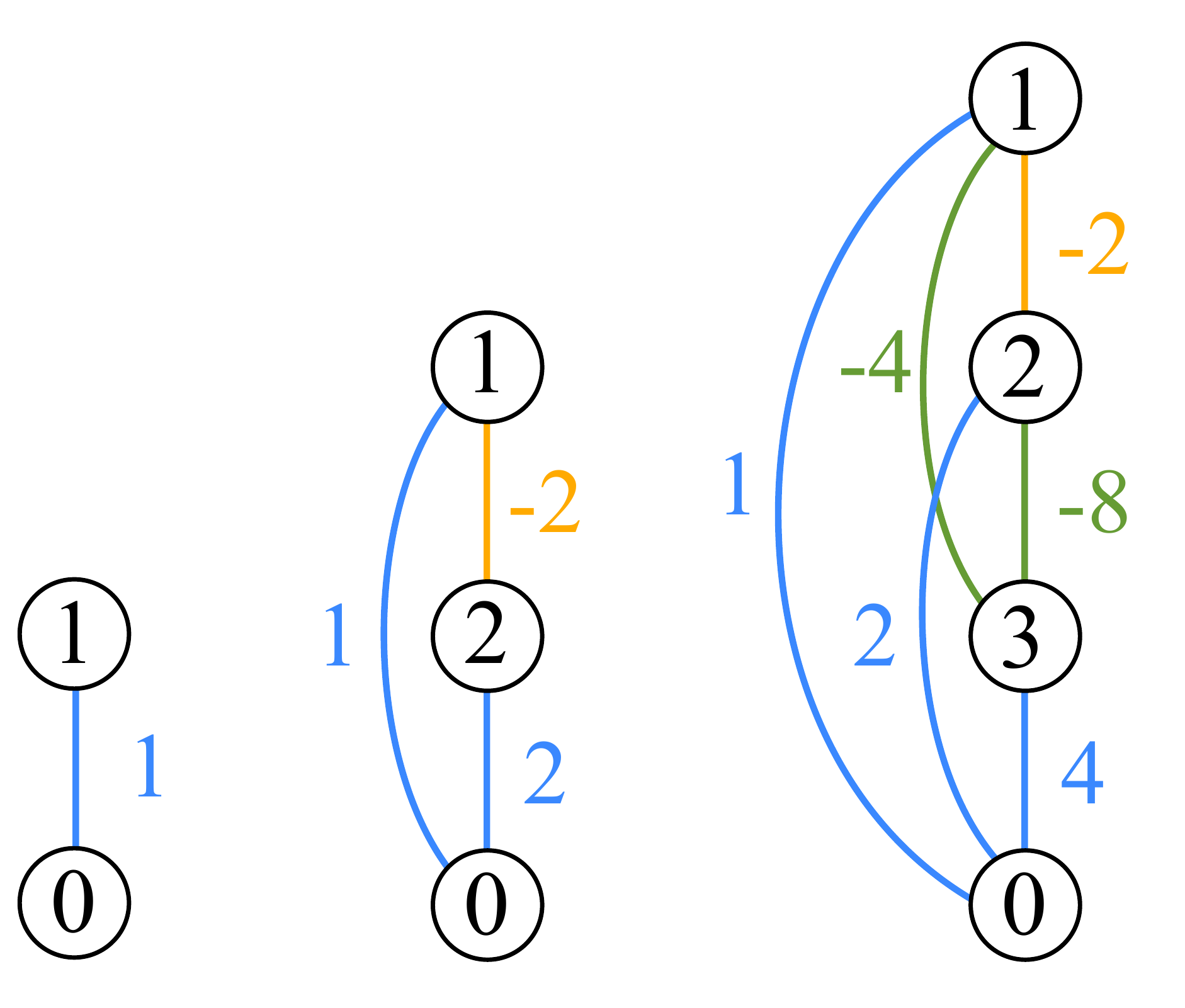}\\
  {\hspace{4mm}(a)\vspace{2mm}}
  \includegraphics[width=0.9\textwidth]{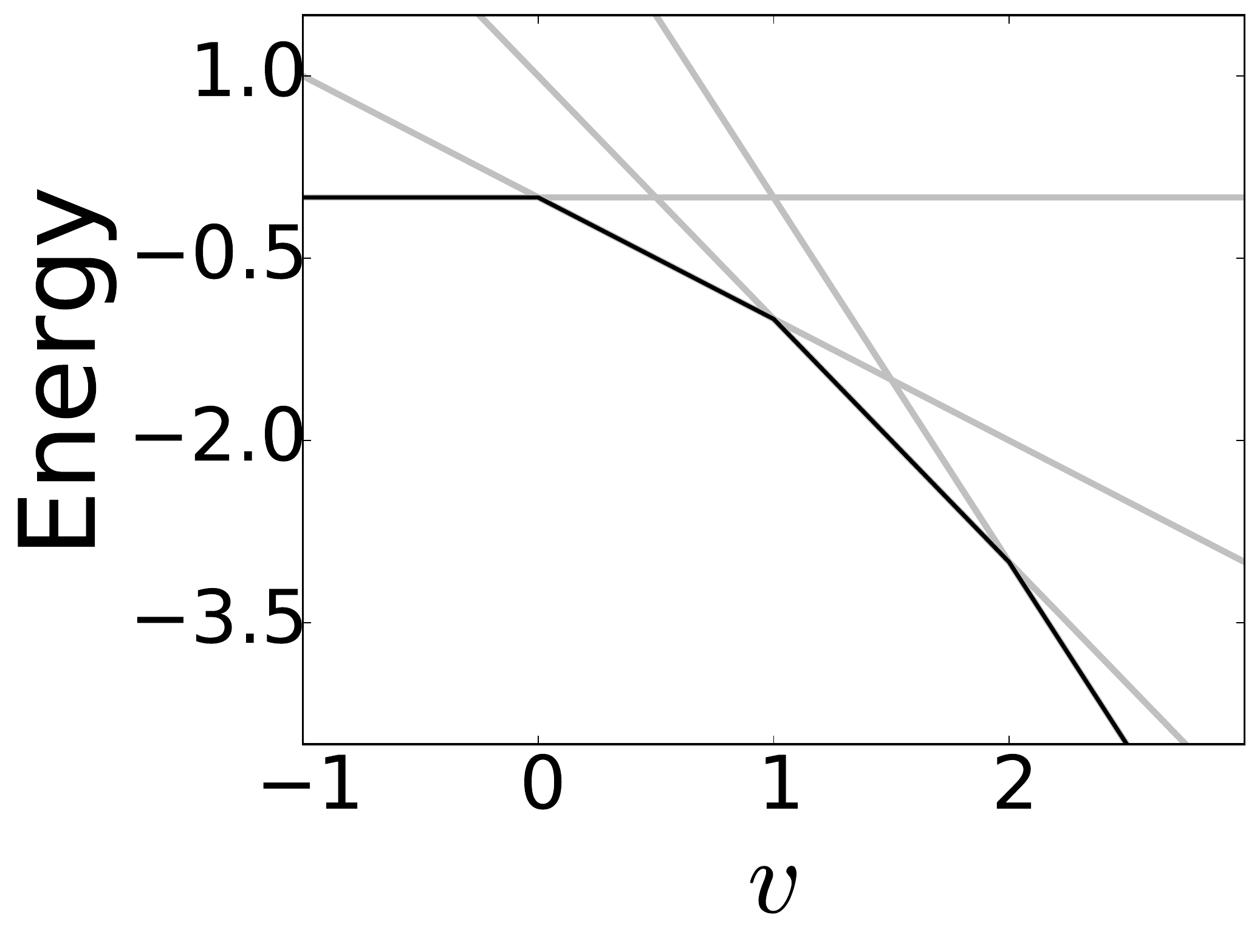}\\
  {\hspace{4mm}(c)}
 \end{minipage}
 \begin{minipage}{0.49\textwidth}
  \centering
  \includegraphics[width=0.9\textwidth]{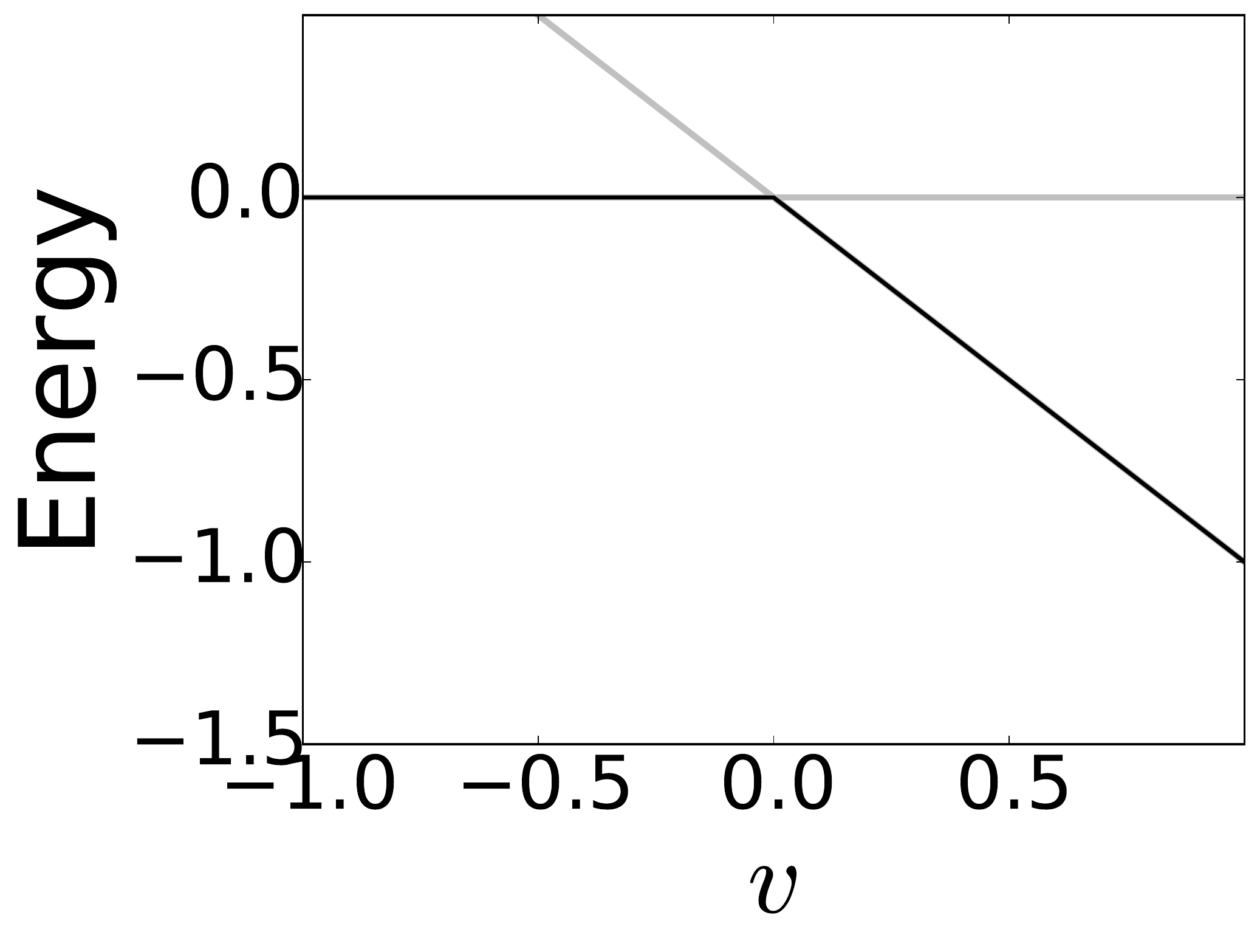}\\
  \hspace{4mm}(b)\vspace{2mm}
  \includegraphics[width=0.96\textwidth]{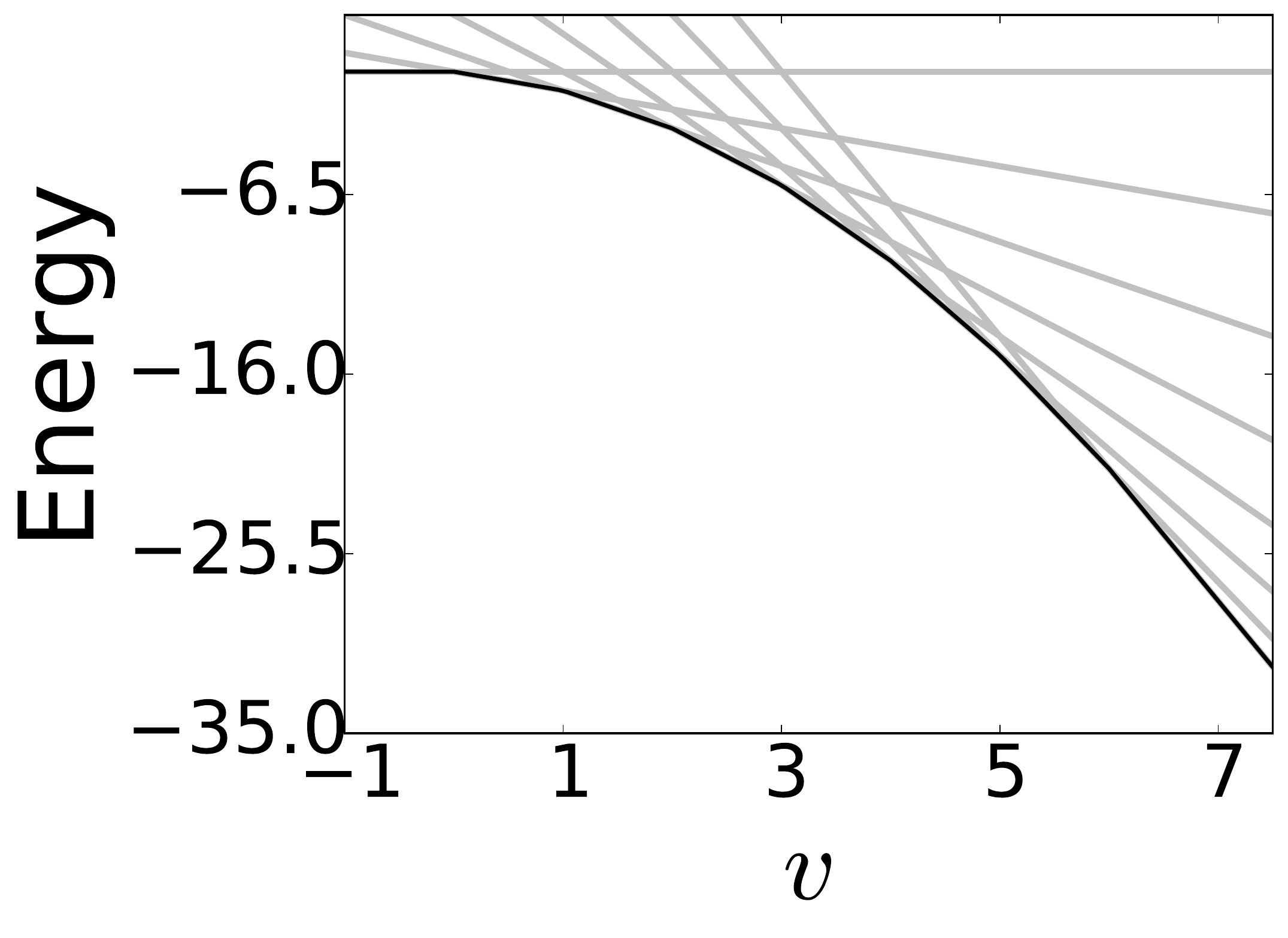}\\
  \hspace{4mm}(d)
 \end{minipage}
\caption{{\bf (a)}: \fBM[L]\ for $L\in\{1,2,3\}$ with unit indices and connection strengths.
{\bf (b) to (d)}: $\hmf$ (printed in black) and $E(v,\bullet)\in\Sl$ (in gray) with $L=$ (b)1, (c)2, and (d)3. }
\label{gBMl_fe:fig}
\end{minipage}
 \end{figure}


\begin{wrapfigure}{L}{0.32\textwidth}
 \includegraphics[width=0.3\textwidth]{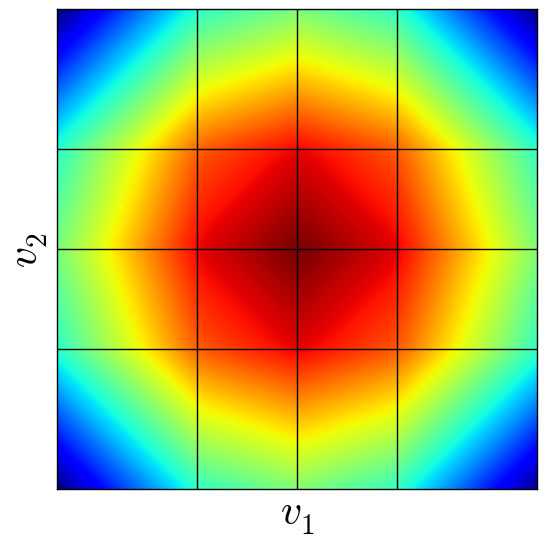}
\caption{Heatmap of $F$ of a two-layered sDBM. 
The sDBM is constructed as a bundle of two independent gBM(2)s. 
Lines indicate the boundary of linear regions of $\hmf$. 
The parameters are computed with Algorithm~\ref{mb1DBM:alg} and with rescaling. 
}\vspace{-7mm}
 \label{sDBM_fe2d:fig}
\end{wrapfigure}

We first analyze an arrangement of the elements of $\Sl$ with a network construction procedure and then analyze the number of linear regions of $\hmf$ under this arrangement. 
The procedure is listed in Algorithm~\ref{mb1DBM:alg}, where a network is constructed by appending a unit in a recursive manner, starting from the $1\stus$ unit to the $L\thus$ unit (see \reffig{gBMl_fe:fig}~(a) for example). 
With this construction, we can show that all the elements of $\Sl$ are arranged to be a tangent of a quadratic curve at $2^L$ different points:
\begin{lemma}
Assume that $\{\wkl\}, \{\bk\}$ are computed with \Call{softDeep}{$M$} for a large integer $M$. Then elements of $\Sl$ for $0 < L\leq M$ are tangents of a quadratic function with equally spaced points of tangency. \label{tangency:th} 
\end{lemma}
From Lemma~\ref{tangency:th}, we can readily show that:
\begin{lemma}
The number of effective mixtures of a \fBM\ reaches $2^L=2^{\noh}$, the bound in Proposition~\ref{gen_ub:th}, when parameters are properly set. \label{nolr_fBM:th}
\end{lemma}
Figures~\ref{gBMl_fe:fig} from (b) to (d) demonstrate the statements of Lemma~\ref{tangency:th} and \ref{nolr_fBM:th} for $L\in\{1,2,3\}$. 
Different hidden units control the slope of the energy in different levels of magnitude (e.g., $1,2,4,8,\ldots$). 
This allows $E(v,\bullet)$ to have mutually different slopes and thus leads to $2^L$ effective mixtures. 


We call connections determined with Algorithm~\ref{mb1DBM:alg} {\em soft-deep} connections because 
the connection strengths can be regarded to decay exponentially in the distance between layers.
Let us have units aligned in a sequential order $k=0,L,\ldots,1$ as in \reffig{gBMl_fe:fig}~(a). 
The strength of a connection from a unit to an upper unit which is $d$ units away from the unit under this spatial configuration is proportional to $2^{-d}$. 
We observe that this connection pattern is {\em soft} counterpart of the conventional deep connection pattern where only adjacent layers are connected. 




\subsubsection{Main Results}
\label{main_sDBM:sec}
By applying Lemma~\ref{nolr_fBM:th} to an sDBM constructed as a bundle of independent \fBM s, we can show that the maximal number of effective mixtures of an $L$-layered sDBM scales exponentially in $L$:
\begin{theorem}
Suppose an sDBM with $L$ hidden layers each of which contains $M(\leq\nov)$ units. Then the number of effective mixtures of this sDBM reaches $2^{ML}=2^\noh$, the bound in Proposition~\ref{gen_ub:th},  with a certain parameter configuration. 
\label{main:th}
\end{theorem}

\refFig{sDBM_fe2d:fig} demonstrates the claim of Theorem~\ref{main:th}; 
the free energy function of an sDBM with four hidden units can be well approximated with $\hmf$ that has $2^4=16$ linear regions. 

Along with the analysis on DBMs and RBMs, Theorem~\ref{main:th} indicates that soft-deep connections that bypass between remote layers can be {vital} for a deeply layered BM to have superior representational power to shallow one in terms of the efficiency of a distributed representations. This clearly contrasts with feedforward networks where bypassing connections do not critically affect the representational power \cite{Bishop:2006uia}. 


\section{Remarks}



There are two appealing properties of sDBMs other than the huge number of effective mixtures. 
First, fast block Gibbs sampling can be performed. 
Although sampling efficiency degrades compared to DBMs due to the dependency between hidden layers introduced by soft-deep connections, 
we believe that benefits from the huge representational power offset this negative effect. 

Second, soft-deep connections can ease difficulties in learning deeply layered BMs. 
Because DBMs do not have connections between remote layers, the effect that the visible layer exerts on remote hidden layers decays exponentially in the depth. 
This phenomenon will hinder learning signals from correctly propagating through deep layers. 
We believe that one of the benefits of pretraining is to help this stochastic vanishing gradient effect.
The soft-deep connections ease this problem by bypassing between the visible layer and remote hidden layers. 
We believe that high generative performance of sDBMs without pretraining shown in Section~\ref{exp:sec} is achieved not only with the huge representational power proven in Theorem~\ref{main:th}, but also with the less severe vanishing gradient effect. 


\subsection{Soft-Deep Regularization}
For the number of effective mixtures of an sDBM to scale exponentially in the depth as in Theorem~\ref{main:th}, it is essential that connection weights are in multiple levels of magnitude. 
Without regularization or with uniform regularization, networks do not attain such property via learning. 
To address this point, we introduce soft-deep regularization where strength of L2 regularization for connections between $k\thus$ and $l\thus$ layers is inversely proportional to $|\wkl|^\eta$ where $\wkl$ are computed with Algorithm~\ref{mb1DBM:alg}, and $\eta$ is a hyper parameter. 
Although this technique does not strictly guarantee that the weights scale as $\wkl$, we experimentally observed that this regularization improves the performance of sDBMs. 


\section{Experiments}
\label{exp:sec}
We have discussed representational power of BMs based on approximated free energy $\hmf$.
To validate the approximation, we experimentally demonstrated advantages of sDBMs.
We performed experiments on two datasets: MNIST digits \cite{Lecun:1998hy} and Caltech-101 silhouettes \cite{Marlin:2012ue}. 
We used Theano \cite{Bastien:2012wr} and pylearn2 \cite{Goodfellow:2013wv} to implement sDBMs. 
We used stochastic maximum likelihood \cite{Tieleman:2008gw} to jointly train networks with the centering method \cite{Anonymous:88gzkJva} and soft-deep regularization. We did not perform pretraining. 
We scheduled learning rates to linearly decay from an initial value to zero. 

To evaluate networks, we used AIS \cite{Neal:2001wea, Salakhutdinov:2008eq} to estimate the variational lower bound for the average log-likelihood on test data.
We evaluated the reliability of estimates by computing $3\sigma$ confidence intervals, which we show in the supplementary material. 

On both datasets, we trained 2-, 3-, and 4-layered sDBMs with various hyper parameters. 
The number of units in each hidden layer is fixed to 500.
Hyper parameters are tuned via random sampling \cite{Bergstra:2012rs}.
See supplementary material for detail. 


\begin{figure}[h]
\centering
\includegraphics[width=1.0\linewidth]{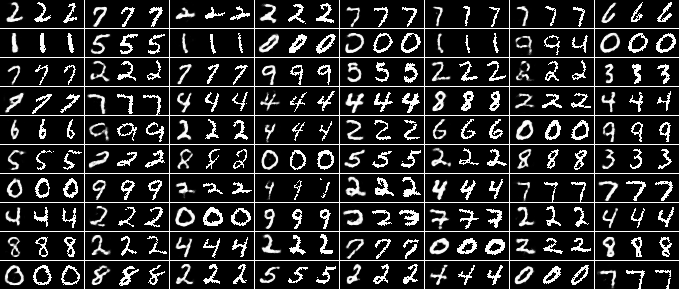}
\vspace{-5mm}
 \caption{
Random samples from a 4-layered sDBM (left in each cell) 
displayed with nearest training (center) and test samples (right) from binarized MNIST. 
Generated images are probabilities that pixels are sampled from. 
The nearest neighbors are computed in terms of pixelwise $L^2$ distance. 
The sDBM does not simply memorize training examples but generalize to unseen test examples.  
}
\label{MNIST_sample:fig}
\end{figure}

\begin{figure}[h]
\centering
\includegraphics[width=1.0\linewidth]{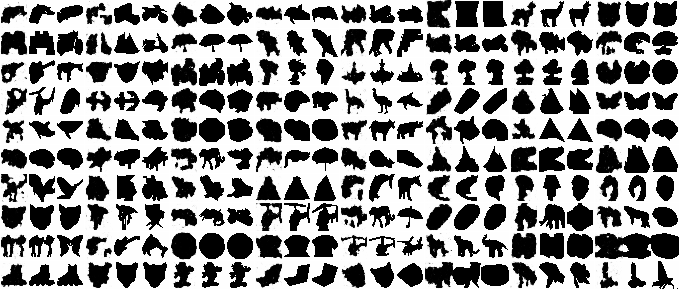}
\vspace{-5mm}
 \caption{Random samples from a 4-layered sDBM trained on Caltech-101 silhouettes displayed with 
nearest training and test examples as in \reffig{MNIST_sample:fig}.}
\label{SIL_sample:fig}
\end{figure}

\subsection{Binarized MNIST}
MNIST is a collection of gray scaled digit images that consists of 60,000 training samples and 10,000 test samples \cite{Lecun:1998hy}. We binarized the images following the procedure by \citet{Salakhutdinov:2008eq} to generate training and test data. 



\reftab{MNIST:tab} compares sDBMs and various models in the literature in terms of generative performance. 
We can see that sDBMs greatly performed compared to other models. 
Even the 2-layered sDBM outperformed the previous state-of-the-art test log-likelihood of $-80.97$ nat by a recent report \cite{Gregor:2015up}. 
The best-performing 4-layered sDBM achieved $-66.56$ nat of test log-likelihood with a $3\sigma$ confidence interval of $[-67.01, -65.70]$ nat. 

Note that sDBMs with 2- and 3-hidden layers outperformed DBMs with the same number of layers \cite{Salakhutdinov:2012vo}. 
This result would be seen to reflect the exponentially greater number of effective mixtures of sDBMs 
than of DBMs with the same number of parameters.

The depth of networks largely improved the performance, 
though improvement of the 3-layered model over the 2-layered model was relative small. 
We believe that this effect is due to insufficiency in parameter tuning; 
The 3-layered model performed worse than 2-layered model on training data 
as shown in the supplementary material.
The performance of models would uniformly improve as the depth of networks with more precise parameter tuning. 

\refFig{MNIST_sample:fig} shows random samples from the best performing sDBM.
The sDBM well generalizes to unseen test examples. 


\begin{table}[h]
\newcolumntype{R}{>{\raggedleft\arraybackslash}X}
\newcolumntype{L}{>{\raggedright\arraybackslash}X}
\caption{Comparison of generative performance of various generative models on binarized MNIST. We report average test log-likelihood measured in nat. 
}
\label{MNIST:tab}

\begin{tabularx}{\columnwidth}[t]{|L|cc||L|cc|}\hline
Model & Test LL & $\geq$ &Model& Test LL & $\geq$\\ \hline \hline
RBM \cite{Salakhutdinov:2008eq}&$\approx$ -86.34&&DLGM 8 leapfrogs\cite{Salimans:2014tn}&$\approx$ -85.51&-88.30\\
DBN 2hl \cite{Murray:2008to}&$\approx$ -84.55&&DARN 1hl \cite{Gregor:2014ti}&$\approx$ -84.13&-88.30\\
DBM 2hl \cite{Salakhutdinov:2012vo}&$\approx$ -83.43&&DARN 12hl \cite{Gregor:2014ti}&--&-87.72\\
DBM 3hl \cite{Salakhutdinov:2012vo}&$\approx$ -83.02&&DRAW \cite{Gregor:2015up}&--&{ -80.97}\\ \cline{4-6}
NADE \cite{Larochelle:2011wd}&-88.33&&sDBM 2hl&--&{\bf -76.41}\\
EoNADE 2hl \cite{Uria:2014tu}&-85.10&&sDBM 3hl&--&{\bf -74.58}\\
EoNADE-5 2hl \cite{Raiko:2014wa}&-84.68&&sDBM 4hl&--&{\bf -66.56}\\
DLGM \cite{Rezende:2014vm}&$\approx$ -86.60&&&&\\
\hline
\end{tabularx}

\end{table}

\subsection{Caltech-101 silhouettes}

Caltech-101 silhouettes is a collection of binary silhouette images of various objects \cite{Marlin:2012ue}. The dataset contains 4,100 training samples and 2,307 test samples. 

\reftab{silhouettes:tab} compares sDBMs with several other models on generation of Caltech-101 silhouettes. 
sDBMs outperformed the previous state-of-the-art by NADE trained with reweighted wake-sleep (RWS) algorithm \cite{Bornschein:2014uj}. 
The 4-layered sDBM achieved $-85.55$ nat of test log-likelihood with a $3\sigma$ confidence interval of $[-85.67, -85.40]$ nat. 
This result is the best average test log-likelihood achieved on Caltech-101 silhouettes to the best of our knowledge. 


The depth improves the performance of sDBMs. 
We believe that less precise parameter tuning resulted in
poor performance of 2-layered model as in experiments with MNIST. 

\refFig{SIL_sample:fig} shows samples generated from the best performing 4-layered model. 
The most samples proves nice generalization by the network though some samples resemble training examples.




\begin{table}
\newcolumntype{R}{>{\raggedleft\arraybackslash}X}
\newcolumntype{L}{>{\raggedright\arraybackslash}X}
\caption{Comparison of generative performance of various generative models on Caltech-101 silhouettes. 
We report average test log-likelihood as in \reftab{MNIST:tab}. 
}
\label{silhouettes:tab}
\begin{tabularx}{\columnwidth}[t]{|Lr|c||Lr|c|}\hline
Model&& Test LL & Model&& Test LL\\ \hline \hline
RBM \cite{Cho:2013iea}&&-109.0&NADE-RWS&&-104.3\\ \cline{4-6}
RBM \cite{Cho:2013iea}&&-107.8&sDBM 2hl&&$\geq${\bf -92.4}\\
NADE-2 \cite{Raiko:2014wa}&&-108.8&sDBM 3hl&&$\geq${\bf -98.7}\\
NADE-5\cite{Raiko:2014wa}&&-107.8&sDBM 4hl&&$\geq${\bf -85.5}\\
\hline
\end{tabularx}

\end{table}

\section{Conclusion}

In this paper, we proposed a BM architecture that can better exploit the distributed representation. 
We proposed a measure for the efficiency of a distributed representation of a BM, the number of effective mixtures of a BM, which is the number of linear regions of a piecewise linear function that approximates a free energy function of a BM. 
We showed inefficiency of DBMs with respect to the maximal number of effective mixtures. 
We proposed sDBMs, an extension of DBMs. 
We showed that the maximal number of effective mixtures of an sDBM is exponentially larger than that of a RBM or a DBM. 
Finally, we experimentally demonstrated high generative performance of sDBMs.

 \section*{Acknowledgement}
 This research was supported by JSPS Grant-in-Aid for JSPS Fellows (145500000159).  We thank KyungHyun Cho, Li Yao, Takaki Makino, and Keita Tokuda for valuable discussion. 

\bibliographystyle{unsrtnat}
\bibliography{nips2015}
\newpage
\appendix

\section{Boltzmann Machines}
A Boltzmann machine (BM) is a stochastic generative model, which is typically defined over $N$ binary units $\xj[i]\in\binary$. 
The probability that a BM assigns to a state $\xs=\{\xj[i]\}$ is defined as
\begin{align}
p(\xs;\para) &= \frac{1}{\partf}\exp(-\ebm(\xs;\para)), \label{prob:eq}
\end{align}
where the normalization constant, or the partition function of the BM is denoted by $\partf$, and its energy function is defined as 
\begin{align}
\ebm(\xs;\para) &= - \sum_{i=1}^{N}\sum_{j=1}^{N}\xj[i] w_{i,j}\xj - \sum_{i=1}^{N}b_i\xj[i]\label{bm:eq},
s\end{align}
where $w_{i,j}$ are symmetric (i.e., $w_{i,j}=w_{j,i}, w_{i,i}=0$) connection weights between units $i$ and $j$, $b_i$ are biases, and $\para$ is the set of the parameters. 

A BM with visible and hidden units can have rich representations; visible units $\vj\in\vv$ correspond to data variables, and hidden units $\hi\in\hs$ correspond to latent features of data. All the units are either a visible units or a hidden unit (i.e., $\xs=\{\hs, \vv\}$). The numbers of visible and hidden units are denoted by $\nov$ and $\noh$. 


Various network topologies that restrict connections of BMs have attracted great research interests \cite{Salakhutdinov:2009uo, Rumelhart:va, osindero:2008mo, srivastava:2012mu}. 
Albeit general BMs are a superset of such BMs with restricted connections, general BMs are rarely used in practice. 
The main problem with general BMs is the difficulty due to the intractability of the expectations with respect to data-dependent and model distributions. One approach is to approximate expectations via expensive MCMC \cite{hinton:1983tf}. The relaxation time of a Markov chain can be quite long because general BMs have enormous number of well-separated modes to be explored. Moreover, dense connections of BMs require generic Gibbs sampling which updates only one unit at a time. 
Restriction on connections can alleviate these issues. 
Particularly, BMs with layered connection patterns are widely studied because of their appealing properties such as efficiency in sampling, less-complex energy landscapes, and simplicity of learning algorithms. 
We here review two representative layered BMs: Restricted BMs (RBMs) and Deep BMs (DBMs).  

\subsection{RBMs}
An RBM is a BM with a bipartite graph that consists of a visible layer and a hidden layer. Connections within each layer are restricted \cite{Rumelhart:va}. The energy function of an RBM is defined as 
\begin{align}
\ebm(\{\hvk[1], \vv\};\prbm) = -\sum_{i=1}^{\nn[1]}\sum_{j=1}^{\nn[0]} \hik[i][1] \wijkl[i][j][1][0] \vj - \sum_{j=1}^{\nn[0]} \bik[j][0]\vj - \sum_{i=1}^{\nn[1]} \bik[i][1]\hik[i][1],\label{rbm:eq}
\end{align}
where $\hvk[1]$ denotes the states of the (first) hidden layer, and $\prbm = \{\wijkl[i][j][1][0], \bik[j][0], \bik[i][1]\}$ are model parameters. 
We here use redundant notation with layer indices associated with a superscript to avoid confusion of notations for models which we shall describe in later sections. 

RBMs exhibit a nice property that conditional distributions $p(\hvk[1]|\vv)$ and $p(\vv|\hvk[1])$ are tractable and factorized. 
This allows us to perform fast block Gibbs sampling and makes the data-dependent expectation tractable. 

However, such tractability substantially sacrifices the representation power of RBMs. 

\subsection{DBMs}

DBMs are an extension of RBMs that have multiple hidden layers that form deep hierarchy. 
Connections within each layer are restricted, and units in a layer are connected to all the units in the neighboring layers \cite{Salakhutdinov:2009uo}. The energy function of a DBM with $L$ layers is 
\begin{align}
\ebm(\xs;\pdbm) = -\sum_{k=0}^{L-1}\sum_{i=1}^{\nn[k+1]}\sum_{j=1}^{\nn} \xik[i][k+1] \wijkl[i][j][k+1][k] \xik[j][k] - \sum_{k=0}^{L}\sum_{i=1}^{\nn[k]} \bik\xik,
\label{dbm:eq}
\end{align}
where we number layers s.t. the 0th layer is the visible layer, and the $k\thus$ layer is the $k\thus$ hidden layer. 
The state of the $k\thus$ layer is denoted by $\xvk=\{\xik\}$, hence the state of the $k\thus$ hidden layer is $\hvk = \xvk,\,(\hik=\xik)$ for $0<k\leq L$, and the state of the visible layer is $\vv = \xvk[0],\,(\vj=\xik[j][0])$. 
Let $\nn$ be the number of units in $k\thus$ layer (i.e., $\nov=\nn[0]$, $\noh = \sum_{k=1}^L\nn$). 


DBMs have several appealing properties. 
Fast block Gibbs sampling is also applicable to DBMs, as to RBMs. 
Particularly, block sampling is highly efficient because conditional distributions of the even layers given the odd layers and those of the odd layers given the even layers are tractable and factorized. 
Moreover, DBMs possess greater representation power than RBMs because of multiple hidden layers. 

However, the improved representation power causes a serious difficulty. The data-dependent expectation  needs to be approximated in learning because the conditional distribution $p(\hs|\vv)$ is no longer tractable; stochastic approximation procedure \cite{Tieleman:2008gw} or variational inference \cite{Salakhutdinov:2009uo, Salakhutdinov:2010vo} is used for approximation.
At the appearance of DBMs, \citet{Salakhutdinov:2009uo} introduced a pre-training algorithm to ease this problem. 
Recently, the centering method is proposed for joint training of DBMs without pre-training \cite{Anonymous:88gzkJva}.

Upon the introduction of DBMs, DBMs would have been expected to be scalable, i.e., great performance improvements can be achieved with DBMs by stacking a layer as in other deep neural models. 
However, experiments suggest that this seems not true; improvements are hard to be gained with {\it very deep} BMs with more than 3 hidden layers even with elaborated learning algorithms \cite{Salakhutdinov:2009uo}. 
It is widely conceived that the poor scalability of DBMs is attributed that we cannot exploit huge representation capacity of DBMs due to inefficient optimization methods. This will be true to some extent.
We, however, shall provide both empirical and theoretical evidences that the poor scalability of DBMs is not only due to the optimization issues, but also because of rather limited representation capacity of DBMs.



\section{Proof of Theorems}

\setcounter{theorem}{1}
\begin{theorem}
 Let $\eres(\vv) = - \log\{\sum_\hs \exp(-E(\vv,\hs))  - \exp(-\hmf(\vv))\}$. Then the free energy $\fbm(\vv)$ is bounded as:
\begin{align}
 \hmf(\vv) - \exp(\hmf(\vv)-\eres(\vv))\leq \fbm(\vv)\leq \varf(\vv) \leq\hmf(\vv), 
\end{align}
where $\varf$ is the mean-filed approximation of the free energy. 

\begin{proof}
We first show the upper bound. For any approximating posterior $Q(\hs|\vv)$, We have
\begin{align}
 \fbm(\vv) &= \sum_\hs Q(\hs|\vv) E(\hs, \vv) - \sum_{\hs}Q(\hs|\vv)\log\frac{1}{Q(\hs|\vv)} - \KL{Q(\bullet|\vv)}{P(\bullet|\vv)},
 \label{fapprox:eq}
\end{align}
where $P(\hs|\vv)$ denotes the model's true posterior distribution. Suppose we have a following appromximating posterior:
\begin{align}
 \qmin(\hs|\vv) = 
\left\{
\begin{array}{cl}
 1&(\hs=\hsmin)\\
 0&(otherwise)\\
\end{array}\right.,
\end{align}
where we defined $\hsmin=\argmin_{\hs} E(\vv, \hs)$. Note that this posterior factorizes. With this posterior, \refeq{fapprox:eq} becomes
\begin{align}
 \fbm(\vv) &= \hmf(\vv) -  \KL{\qmin(\bullet|\vv)}{P(\bullet|\vv)}.
\end{align}
we have $\fbm(\vv) \leq \hmf(\vv)$ because $\KL{\qmin(\bullet|\vv)}{P(\bullet|\vv)} = -\log P(\hsmin|\vv) \geq 0$. 
The equality holds if and only if the true posterior has its all the mass on $\hsmin$, i.e., $P(\hs|\vv)=\qmin(\hs|\vv)$. 
We have an inequality $\varf(\vv) \leq\hmf(\vv)$ readily from the definition of the mean-field free energy because $\qmin$ is factorized. 

Next, we prove the lower bound as:
\begin{align}
 \hmf(\vv) - \fbm(\vv) &= \hmf(\vv) + \log\sum_{\hs} \exp(-E(\{\hs, \vv\})) \\
 &= \log\left(\frac{\exp(-\hmf(\vv)) + \exp(-\eres(\vv))}{\exp(-\hmf(\vv))}\right)\\
 &=  \log\left(1 + \exp(-\eres(\vv) + \hmf(\vv))\right)\\
 & \leq  \exp(\hmf(\vv)-\eres(\vv)), 
\end{align}
where we used $\log(x)\leq x-1$ in the last line. The equality holds if and only if $\hmf(\vv) = \eres(\vv)$. 
\end{proof}
\end{theorem}

\subsection{On The Number of Effective Mixtures of an RBM}
\setcounter{theorem}{4}

\begin{theorem}
 The maximal number of effective mixtures of an RBM is $\sum_{j=0}^{\nn[0]} {\nn[1]\choose{j}}$.
\begin{proof}
 The hard-min free energy of an RBM can be written as $\hmf(\vv;\prbm) = - \sum_{j=1}^{\nn[0]}\bik[j][0]\vj- \sum_{i=1}^{\nn[1]}\max(0, \sum_{j=1}^{\nn[0]}\wijkl[i][j][1][0]\vj + \bik[i][1]))$. The number of linear regions of this function is the number of regions separated by $\nn[1]$ hyper-planes each of them satisfies $\sum_{j=1}^{\nn[0]}\wijkl[i][j][1][0]\vj + \bik[i][1] = 0$ for $0< i \leq\nn[1]$. The number of these regions is $\sum_{j=0}^{\nn[0]} {\nn[1]\choose{j}}$ \cite{Zaslavsky:1975vb}. This proves the claim.
\end{proof}
\end{theorem}

\subsection{On The Number of Effective Mixtures of a DBM}
Here we provide lower and upper bounds for the maximal number of effective mixtures of DBMs with respect to the parameters. Let us begin with a lower bound. Because DBMs are a superset of RBMs, the number of effective mixtures of a DBM can be as large as the maximal number of effective mixtures of RBMs. This observation leads us to a lower bound: 
\begin{proposition}
The maximal number of effective mixtures of a DBM is lower bounded by $\sum_{j=0}^{\nn[0]} {\nn[1]\choose{j}}$. 
\label{nolr_lb_dbm:th}
\end{proposition}

\setcounter{theorem}{6}
We here outline the idea of the proof of Theorem~\ref{nolr_ub_dbm:th}. 
A key observation is that the marginal distribution over visible units of a DBM is written as a summation: $p(\vv;\pdbm) = \sum_{\hvk[1]}p(\vv|\hvk[1];\pdbm) p(\hvk[1];\pdbm)$ where $p(\hvk[1];\pdbm)$ is the marginal distribution over $\hvk[1]$. 
This indicates that the number of the mixing components of a DBM is $2^{\nn[1]}$; 
this bounds the maximal number of effective mixtures from above . These observations lead to a natural but somewhat shocking result where the bound only depends on the number of units in the first hidden layer:
\begin{theorem}
The number of effective mixtures of a DBM with any number of hidden layers is upper bounded by $2^{\nn[1]}$. 
\begin{proof}
 Suppose a set of linear functions $S(\hvk[1])=\{\ebm(\{\hvk[L], \ldots, \hvk[2], \hvk[1], \vv\})| \hvk[k]\in{\binary}^{\nn[k]} \text{for } 2\leq k\leq L\}$. Linear functions within this set $f\in S(\hvk[1])$ have an identical gradient as $f(v;\hvk[1]) = \sum_{j=1}^{\nn[0]} \alpha_j(\hvk[1]) \vj + C$ where $C$ is a constant that only depends on $\{\hvk[2], \ldots, \hvk[L]\}$ and $\alpha_j(\hvk[1]) = -\sum_{i=1}^{\nn[1]}\hik[i][1] \wijkl[i][j][1][0] - \bik[j][0]$. Therefore, $\min S(\hvk[1])$ is a linear function $f_\mathrm{min}(\vv;\hvk[1]) = \sum_{j=1}^{\nn[0]} \alpha_j(\hvk[1]) \vj + C_\mathrm{min}$ with $C_\mathrm{min} = \min_{\hvk[2], \ldots, \hvk[L]}C$. The hard-min free energy of a DBM is $\hmf(\vv;\pdbm) = \min_{\hvk[1]} \min S(\hvk[1]) = \min_{\hvk[1]}f_\mathrm{min}(\vv;\hvk[1])$, and its maximal number of linear regions is bounded above by the number of configurations of $\hvk[1]$ i.e., $2^{\nn[1]}$. 
\end{proof}
\end{theorem}

These results depict a serious limitation on the representation power of DBMs. 
There are two ways to increase the number of effective mixtures of a DBM. The first way is to stack layers. However, the number of effective mixtures never become greater than $2^{\nn[1]}$, which is solely determined by $\nn[1]$. Therefore, depth does not largely help the capacity of DBMs measured in the number of effective mixtures. 
The second way is to increase $\nn[1]$. This strategy, however, at least necessitates the presence of second layer units, which does not improve the bound $2^{\nn[1]}$. Otherwise, the DBM is equivalent to an RBM, and its maximal number of effective mixtures is merely $\Theta({\nn[1]}^{\nn[0]})$. Therefore, the number of effective mixtures of a DBM is smaller than the upper bound in Proposition~\ref{gen_ub:th}:
\begin{proposition}
 The number of effective mixture of a DBM with $\nn[1]>\nn[0]$ never achieves the bound $2^\noh$.
\begin{proof}
 First, suppose that the DBM has no hidden layers above the first layer. This DBM is equivalent to an RBM, thus from Theorem~\ref{nolr_rbm:th}, the maximal number of effective mixtures of this DBM is smaller than $2^\noh$.  
Next, suppose that the DBM has more than one hidden units in its third hidden layer with non-zero connection weights between units in the second hidden layer. From Theorem~\ref{nolr_ub_dbm:th}, the number of effective mixtures of this DBM is bounded above by $2^{\nn[1]} < 2^{\noh}$. This proves the claim.
\end{proof}
\end{proposition}

\subsection{On The Number of Effective Mixtures of an sDBM}

\begin{lemma}
Assume that $\{\wkl\}, \{\bk\}$ are computed with \Call{softDeep}{$M$} for a large integer $M$. Then elements of $\Sl$ for $0 < L\leq M$ are tangents of a quadratic function with equally spaced points of tangency. 
\begin{proof}
We here show the claim with induction with a quadratic function
\begin{align}
 f(\xk[0])=-0.5(\xk[0](\xk[0]+1) + 0.25).
\end{align}
As in main text, let $\Sl$ be a set of linear functions $\{E(\xk[0],\xk[1:L])| \xk[1:L]\in\binary^L \}$. 
Assume that elements of $\Sl[L-1]$ are a tangent of $f(\xk[0])$ where the point of tangency is $\xi_{\xk[1:L-1]} = \sum_{k=1}^{L-1} \xk 2^{k-1} - 0.5$, and the slope is $-\sum_{k=1}^{L-1} \xk 2^{k-1}$. 
We divide $\Sl[L]$ into two sets $\Slc[0]$ and $\Slc[1]$, each of which is a set of lines that correspond to either $\xk[L] = 0$ or $\xk[L] = 1$. 
We can readily show that elements of $\Slc[0]$ are a tangent of $f(\xk[0])$ because $\Sl[L-1] = \Slc[0]$. 

We can show the tangency of elements of $\Slc[1]$ as follows. 
Let $g_{\xk[L]=\eta}(\xk[0];\xk[1:L-1])$ be an element of $\Slc[\eta]$ with hidden configuration $\xk[1:L-1]$ for $\eta\in\binary$, i.e., 
\begin{align}
 g_{\xk[L]=\eta}(\xk[0];\xk[1:L-1]) = \ebm(\xk[0],\ldots, \xk[L];\pmbOnedbm[(L)])|_{\xk[L]=\eta}.
\end{align}
Let us consider the difference $g_{\xk[L]=1}(\xk[0];\xk[1:L-1]) - g_{\xk[L]=0}(\xk[0];\xk[1:L-1]) = B_{\xk[1:L-1]}\xk[0] + C_{\xk[1:L-1]}$ where $B_{\xk[1:L-1]}$ and $C_{\xk[1:L-1]}$ can be computed as follows: 
\begin{align}
 B_{\xk[1:L-1]} = -\wkl[L][0] = -2^{L-1}, \label{B:eq}
\end{align} 
and
\begin{align}
C_{\xk[1:L-1]} &=-\sum_{0< l < L} \wkl[L][l] \xk[l] - \bk[L] \\
&= - \wkl[L][0]\sum_{0< l < L} \xk[l]2^{l-1} + 0.5\left((\wkl[L][0])^2 - \wkl[L][0]\right)\\
&=0.5(\wkl[L][0])^2 + \wkl[L][0]\left(\sum_{k=1}^{L-1} \xk 2^{k-1} - 0.5\right)\\
&=0.5(B_{\xk[1:L-1]})^2 + B_{\xk[1:L-1]}\xi_{\xk[1:L-1]}, \label{C:eq}
\end{align} 
where we used $\wkl[L][l] = \wkl[L][0]2^{l-1}$. 

From Eqs.~\ref{B:eq} and \ref{C:eq}, $g_{\xk[L]=1}(\xk[0];\xk[1:L-1])$ is a tangent of $f(x)$ because the difference between y-intercepts $c_{\alpha}$ and $c_{\alpha+\beta}$ of two tangents of a quadratic function $ax^2+bx+c$ with slopes $\alpha$ and $\alpha+\beta$ is calculated as $c_{\alpha+\beta} - c_{\alpha} = -\frac{\beta^2}{4a} - \beta x_{\alpha}$ where $x_{\alpha}$ is the point of tangency of the line with slope $\alpha$. The point of tangency of $g_{\xk[L]=1}(\xk[0];\xk[1:L-1])$ is $\xi_{\xk[1:L-1]} + \wkl[L][0] = \sum_{k=1}^{L} \xk 2^{k-1} - 0.5$ and the slope is $-\sum_{k=1}^{L-1} \xk 2^{k-1} - \wkl[L][0] = - \sum_{k=1}^{L} \xk 2^{k-1}$. 
Therefore, elements of $\Sl$ are a tangent of $f(\xk[0])$ if elements of $\Sl[L-1]$ are a tangent of $f(\xk[0])$. 

Observe that $\Sl[0]$ contains only one element $g(\xk[0])=0$; this is a tangent of $f(\xk[0])$ at the point of tangency $\xk[0]=-0.5$. Therefore, elements of $\Sl$ are a tangent of $f(\xk[0])$ for any $L\leq M$. This proves the claim. 

\end{proof}
\end{lemma}

\begin{lemma}
The number of effective mixtures of a \fBM\ reaches $2^L=2^{\noh}$, the bound in Proposition~\ref{gen_ub:th}, when parameters are properly set.
 \begin{proof}
 Assume a \fBM\ whose parameters are generated with \Call{softDeep}{$L$}. From Lemma~\ref{tangency:th}, an element of $\Sl$ is a tangent of $f(\xk[0])=-0.5(\xk[0](\xk[0]+1) + 0.25)$ at different points. 
Because $f$ is strictly concave, $g(\xk[0])\geq f(\xk[0])$ where $g\in\Sl$ and the equality holds at the point of tangency. 
Therefore, for $\hat{g},g\in\Sl\,(\hat{g}\neq g)$, $\hat{g}(\hat{x})=f(\hat{x})< g(\hat{x})$ where $\hat{x}$ is the point of tangency of $\hat{g}$. 
Thus, at a neighbor of $\hat{x}$, $\hmf(\xk[0]) = \hat{g}(\hat{x})$. 
Because elements of $\Sl$ are tangents of $f(\xk[0])$ at $2^L$ different points,  the number of effective mixtures of the \fBM\ is $2^L$. This proves the claim. 
 \end{proof}
\end{lemma}
This result directly indicates that general BMs with more than one visible units can also achieve the maximal number of effective mixtures $2^{\noh}$ with connection weights determined by our construction procedure where visible-hidden connections are replicated for all the visible units.

\begin{figure}[h]
 \centering
\includegraphics[width=0.5\linewidth]{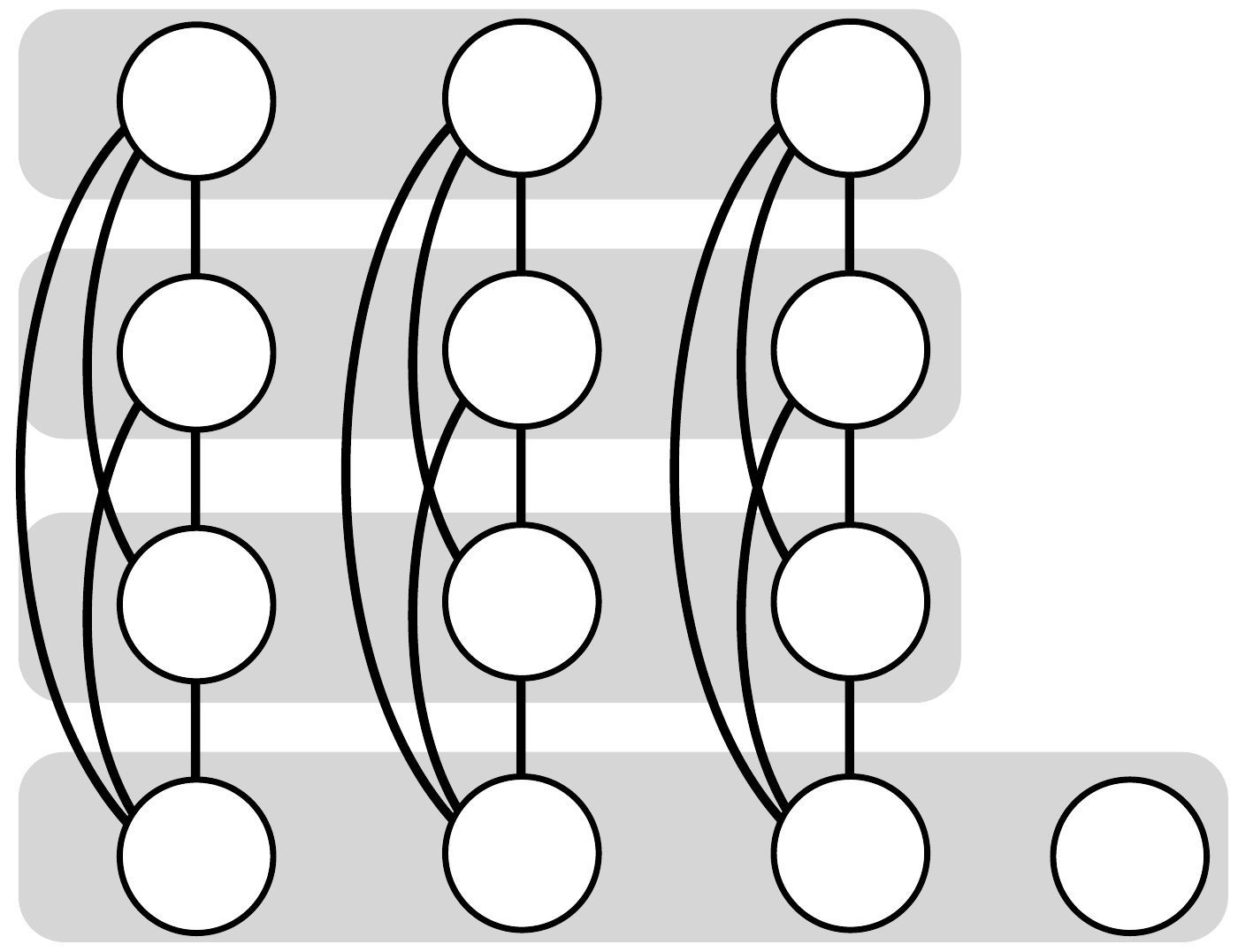} 
\caption{Illustration of an sDBM, which is used in the proof of Theorem~\ref{main:th}. }
\end{figure}

\begin{figure}[h]
\centering
\includegraphics[width=1.0\linewidth]{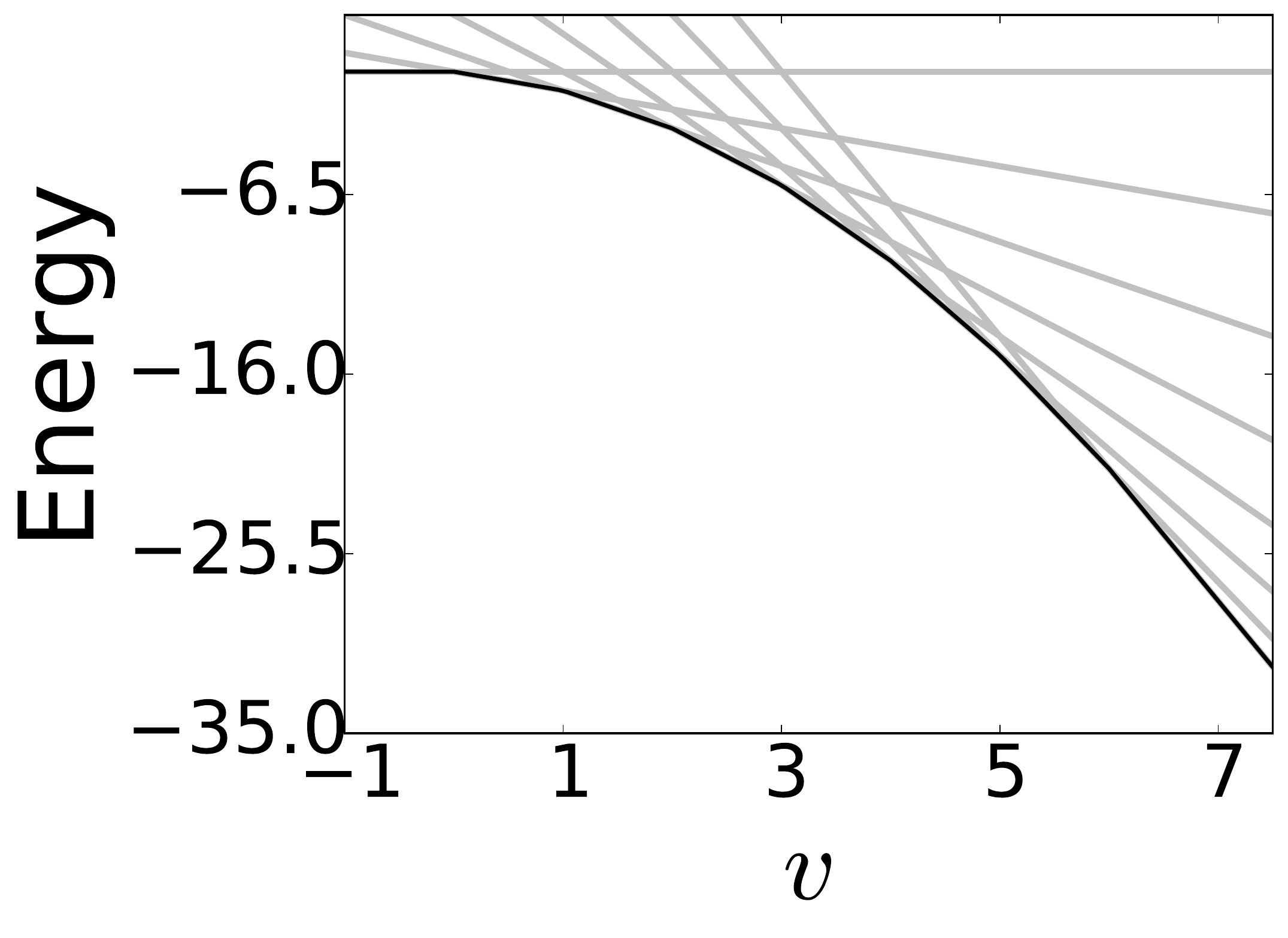}
 \caption{$\hmf$ of \fBM[3] displayed in a large size. $\hmf$ is printed in black, and $E(v,\bullet)\in\Sl$ are in gray.}
\label{large_gBM3_fe:fig}
\end{figure}

\begin{theorem}
Suppose an sDBM with $L$ hidden layers each of which contains $M(\leq\nov)$ units. Then the number of effective mixtures of this sDBM reaches $2^{ML}=2^\noh$, the bound in Proposition~\ref{gen_ub:th},  with a certain parameter configuration. 
\begin{proof}
 Assume an sDBM constructed as a collection of $M$ independent \fBM s each of which has parameters $\pmbOnedbm[(L)]$ generated with \Call{softDeep}{$L$}. 
Then, $\hmf(\vv;\pmbdbm) = \sum_{j=1}^{M} \hmf(\vj;\pmbOnedbm[(L)])$. 
Because each $\hmf(\vj;\pmbOnedbm[(L)])$ has $2^L$ linear regions, the number of linear regions of $\hmf(\vv;\pmbdbm)$ is $2^{ML}$. 
This proves the claim. 
\end{proof}
\end{theorem}

\section{Connection to Biological Neural Nets}

There has been increasingly more intense research interests on the connection between deep neural networks and biological neural networks \cite{Schmidhuber:2014cz}. 
One prevalent aspect is that layers of deep neural networks correspond to cortical regions that form hierarchy \cite{Lee:2007uz}. 
However, unlike conventional deep networks, it is widely known that biological neural networks have many connections that bypass between functionally remote cortical regions (e.g., between V1 and MT) \cite{Felleman:1991tk}. 
Because bypassing connections do not largely contribute to the representation power of feedforward neural networks \cite{Bishop:2006uia}, recent great success of deep feedforward networks do not explain the functional role of such bypassing connections in our brain. 
Our results on sDBMs may help us to understand this mystery.

\section{Details of Experiments}
\subsection{Parameters}
We tuned hyper parameters via random sampling; initial learning rates were sampled from $10^{-[2,4]}$ for MNIST and from $10^{-[2.5,4.5]}$ for Caltech-101 silhouettes, strengths of L2 regularization were sampled from $10^{-[4,7]}$, $\eta$ was sampled from $[0.5,3.5]$, 
and update constants for the centering parameters were sampled from $10^{-[5,8]}$. We generated 16 configurations of hyper parameters for each experiment setting. 
The number of parameter updates was $10^6$. 

Networks were trained with stochastic maximum likelihood \cite{Tieleman:2008gw}. 
We did not perform variational inference \cite{Salakhutdinov:2009uo}. 
The number of positive phase Markov chain updates per parameter update was 5 for MNSIT and 1 for Caltech-101 silhouettes. 
The number of negative phase Markov chain updates per parameter update was 5. 
The batch size was set to 100.

\subsection{AIS}
Throughout the training, we monitored the training and test log-likelihood of models by occasionally performing AIS. 
Such monitoring AIS was executed with rather cheap settings of 100 runs and 30,000 intermediate distributions. 
After training, we performed more expensive AIS on several best performing models evaluated via cheap AIS to gain thorough estimates. 
This expensive AIS is executed with at least 1,000 runs and at least 300,000 intermediate distributions. 
All the figures reported in the main text were gained with such expensive AIS.

\begin{table}
\centering
\newcolumntype{R}{>{\raggedleft\arraybackslash}X}
\newcolumntype{L}{>{\raggedright\arraybackslash}X}
\caption{Details of AIS estimates for sDBMs trained on MNIST. 
Estimated variational lower bounds on training and test data are reported. 
$3\sigma$ confidence intervals are also reported in parentheses. }
\label{mnist_err:tab}
\begin{tabular}[t]{|l|cc|cc|} \hline
Model &\multicolumn{2}{c|}{Train LL} &  \multicolumn{2}{c|}{Test LL}  \\ \hline
sDBM 2hl&-68.80&(-68.91,-68.67)&-76.41&(-76.53,-76.28)\\ 
sDBM 3hl&-71.17&(-71.58,-70.48)&-74.58&(-74.98,-73.89)\\ 
sDBM 4hl&-61.90&(-62.36,-61.04)&-66.56&(-67.01,-65.70)\\ 
\hline
\end{tabular}

\end{table}

\begin{table}
\centering
\newcolumntype{R}{>{\raggedleft\arraybackslash}X}
\newcolumntype{L}{>{\raggedright\arraybackslash}X}
\caption{Details of AIS estimates for sDBMs trained on Caltech-101 silhouttes as in \reftab{mnist_err:tab}}
\label{silhouettes_err:tab}
\begin{tabular}[t]{|l|cc|cc|} \hline
Model &\multicolumn{2}{c|}{Train LL} &  \multicolumn{2}{c|}{Test LL}  \\ \hline
sDBM 2hl&-30.16&(-30.35,-29.92)&-92.37&(-92.56,-92.13)\\ 
sDBM 3hl&-72.62&(-72.69,-72.56)&-98.66&(-98.72,-98.59)\\ 
sDBM 4hl&-38.16&(-38.29,-38.02)&-85.55&(-85.67,-85.40)\\ 
\hline
\end{tabular}

\end{table}

\subsection{Samples from sDBMs}
\begin{figure}[h]
\centering
\includegraphics[width=1.0\linewidth]{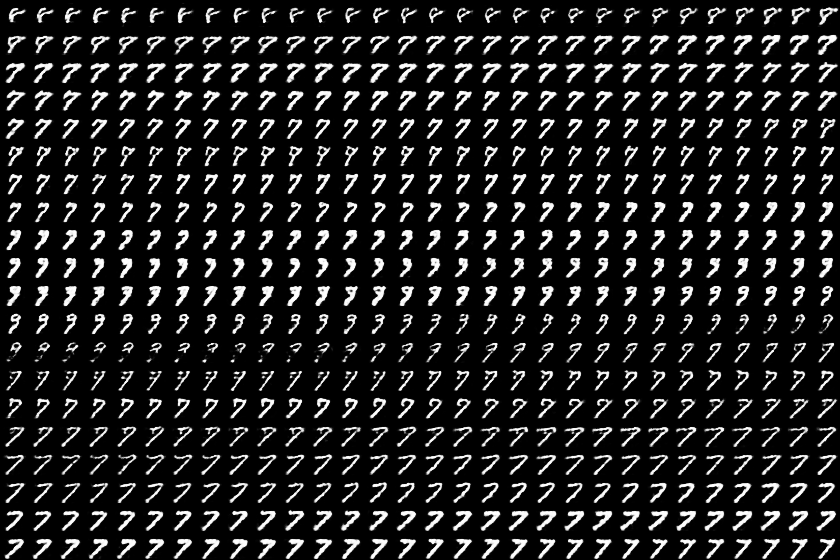}
 \caption{Consecutive samples generated from a 4-layered sDBM trained on MNIST.}
\label{sample_MNIST:fig}
\end{figure}

\begin{figure}[h]
\centering
\includegraphics[width=1.0\linewidth]{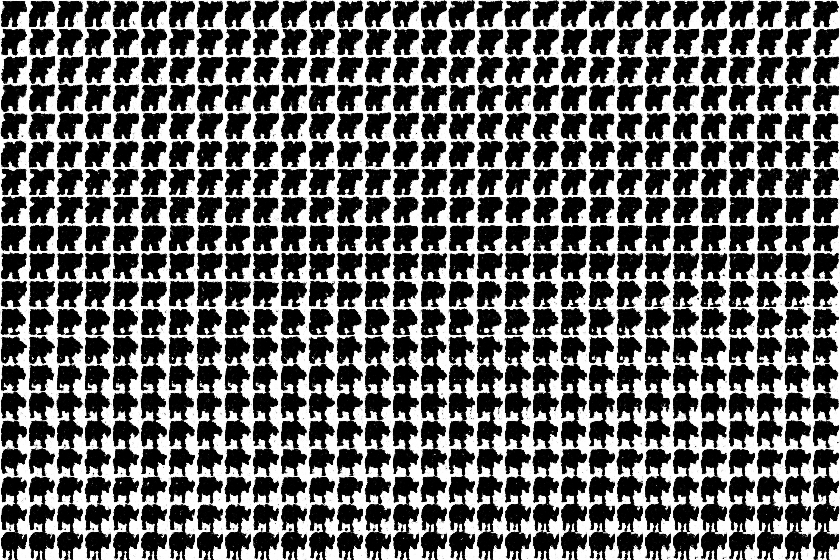}
 \caption{Consecutive samples generated from a 4-layered sDBM trained on Caltech-101 silhouettes.}
\label{sample_SIL:fig}
\end{figure}

Figures~\ref{sample_MNIST:fig} and \ref{sample_SIL:fig} show consecutive samples from the best-performing 4-layered sDBMs.
These figures demonstrate nice mixing of Markov chains between several classes. 

\end{document}